\documentclass[10pt,twocolumn,letterpaper]{article}

\usepackage{cvpr}
\usepackage{times}
\usepackage{epsfig}
\usepackage{graphicx}
\usepackage{amsmath}
\usepackage{amssymb}
\usepackage{booktabs}
\usepackage{float}
\usepackage{multirow}
\usepackage[ruled]{algorithm2e}

\newcommand*{\affaddr}[1]{#1} 
\newcommand*{\affmark}[1][*]{\textsuperscript{#1}}
\newcommand*{\email}[1]{\texttt{#1}}

\usepackage[pagebackref=true,breaklinks=true,letterpaper=true,colorlinks,bookmarks=false]{hyperref}
\usepackage{url}
\cvprfinalcopy 

\def\cvprPaperID{37} 

\begin{document}

\title{Borrow from Anywhere: Pseudo Multi-modal Object Detection in Thermal Imagery}

\author{%
Chaitanya Devaguptapu\affmark[1] \hspace{0.5em} Ninad Akolekar\affmark[1] \hspace{0.5em} Manuj M Sharma\affmark[2] \hspace{0.5em} Vineeth N Balasubramanian\affmark[1] \\
\affaddr{\affmark[1]Indian Institute of Technology, Hyderabad, India}\\
\affaddr{\affmark[2]ANURAG, Defense Research and Development Organization, India}\\
\email{\affmark[1]\{cs19mtech11025,cs16btech11024,vineethnb\}@iith.ac.in}\\
\email{\affmark[2]\{m\_sharma\}@anurag.drdo.in}
}

\maketitle

\begin{abstract}
Can we improve detection in the thermal domain by borrowing features from rich domains like visual RGB? In this paper, we propose a `pseudo-multimodal' object detector trained on natural image domain data to help improve the performance of object detection in thermal images. We assume access to a large-scale dataset in the visual RGB domain and relatively smaller dataset (in terms of instances) in the thermal domain, as is common today. We propose the use of well-known image-to-image translation frameworks to generate pseudo-RGB equivalents of a given thermal image and then use a multi-modal architecture for object detection in the thermal image. We show that our framework outperforms existing benchmarks without the explicit need for paired training examples from the two domains. We also show that our framework has the ability to learn with less data from thermal domain when using our approach. Our code and pre-trained models are made available at: \url{https://github.com/tdchaitanya/MMTOD} 
\end{abstract}

\section{Introduction} \label{introduction}

As indicated by the recent fatalities \cite{ntsbreport}, the current sensors in self-driving vehicles with level 2 and level 3 autonomy (lacking thermal imaging) do not adequately detect vehicles and pedestrians. Pedestrians are especially at risk after dark, when 75\% of the 5,987 U.S. pedestrian fatalities occurred in 2016 \cite{HighwaySafety2017}. Thermal sensors perform well in such conditions where autonomy level 2 and level 3 sensor-suite technologies are challenged. As is well-known, thermal IR cameras are relatively more robust to illumination changes, and can thus be useful for deployment both during the day and night. In addition, they are low-cost, non-intrusive and small in size. Consequently, thermal IR cameras have become increasingly popular in applications such as autonomous driving recently, as well as in other mainstream applications such as security and military surveillance operations. Detection and classification of objects in thermal imagery is thus an important problem to be addressed and invested in, to achieve successes that can be translated to deployment of such models in real-world environments.

Although object detection has always remained an important problem in computer vision, most of the efforts have focused on detecting humans and objects in standard RGB imagery. With the advent of Deep Convolutional Neural Networks (CNNs) \cite{NIPS2012_4824}, object detection performance in the RGB domain has been significantly improved using region-based methods, such as the R-CNN \cite{rcnn} and Fast R-CNN \cite{fastrcnn} that use selective search, as well as Faster R-CNN \cite{fasterrcnn} that uses region-proposal networks to identify regions of interest. 
\begin{figure}[]
\centering
   \includegraphics[width=\hsize, scale=0.5]{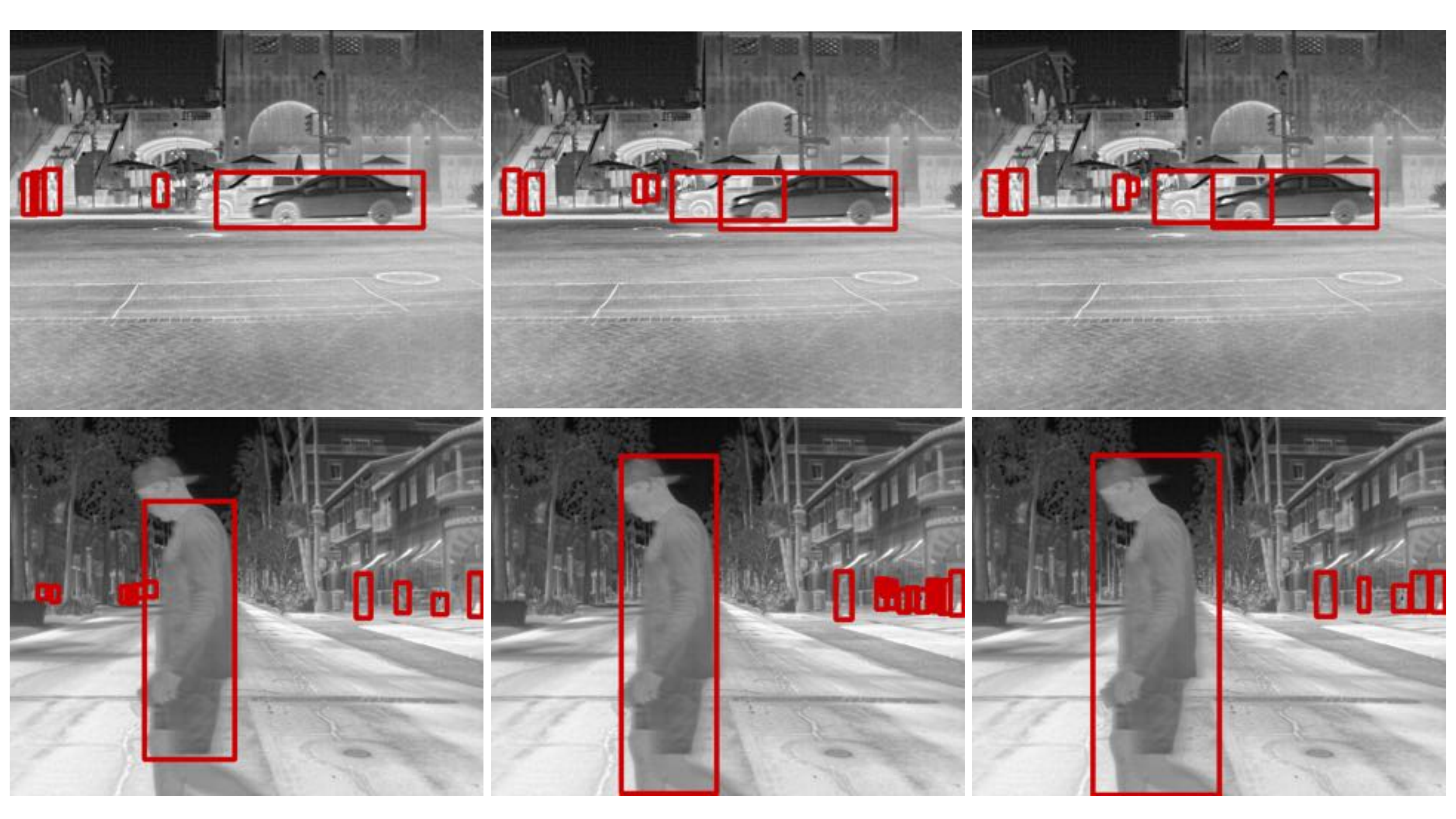}
   \caption{\textit{Left:} Detection with single mode Faster-RCNN; \textit{Middle:} Detection using the proposed method; \textit{Right:} Annotated Ground Truth as provided in FLIR dataset \cite{flir}.}
   \vspace{-1.5em}

\label{fig:teaser}
\end{figure}
Object detection methods such as YOLO \cite{yolo} rephrase the object detection problem into a regression problem, where the coordinates of the bounding boxes and the class probability for each of those boxes are generated simultaneously. This makes YOLO \cite{yolo} extremely fast, although its performance is lower than R-CNN based counterparts \cite{refineDet}.
The aforementioned object detection methods rely, however, on architectures and models that have been trained on large-scale RGB datasets such as ImageNet, PASCAL-VOC, and MS-COCO. A relative dearth of such publicly available large-scale datasets in the thermal domain restricts the achievement of an equivalent level of success of such frameworks on thermal images. 
In this work, we propose a `pseudo multi-modal' framework for object detection in the thermal domain, consisting of two branches. One branch is pre-trained on large-scale RGB datasets (such as PASCAL-VOC or MS-COCO) and finetuned using 
a visual RGB input that is obtained using an image-to-image (I2I) translation framework from a given thermal image (and hence the name `pseudo multi-modal'). The second branch follows the standard training process on a relatively smaller thermal dataset. Our multi-modal architecture helps borrow complex high-level features from the RGB domain to improve object detection in the thermal domain. In particular, our multi-modal approach does not need paired examples from two modalities; our framework can borrow from any large-scale RGB dataset available for object detection and does not need the collection of a synchronized multi-modal dataset. This setting makes this problem challenging too. Our experimental results demonstrate that using our multi-modal framework significantly improves the performance of fully supervised detectors in the thermal domain. The proposed framework also overcomes the problem of inadequacy of training examples in the thermal domain. Furthermore, we also study the relevance of this methodology when there is very limited data in the thermal domain. Our experimental results on the recently released FLIR  ADAS\cite{flir} thermal imagery dataset show that, using only a quarter of the thermal dataset, the proposed multi-modal framework achieves a higher mAP than a single-mode fully-supervised detector trained on the entire dataset. 

The remainder of this paper is organized as follows. Section \ref{relatedwork} provides the context for study including a brief overview of the early and recent work on applying deep learning for thermal imagery. Section \ref{method} describes our approach and methodology. Section \ref{exp_section} describes the experiments carried out and their results. Section \ref{discussion} investigates the impact of size of training set, image resolution and ends with a discussion on some of the cases where our model fails to perform as well.

\begin{figure*}
\begin{center}
\includegraphics[width=16.5cm, height=10cm]{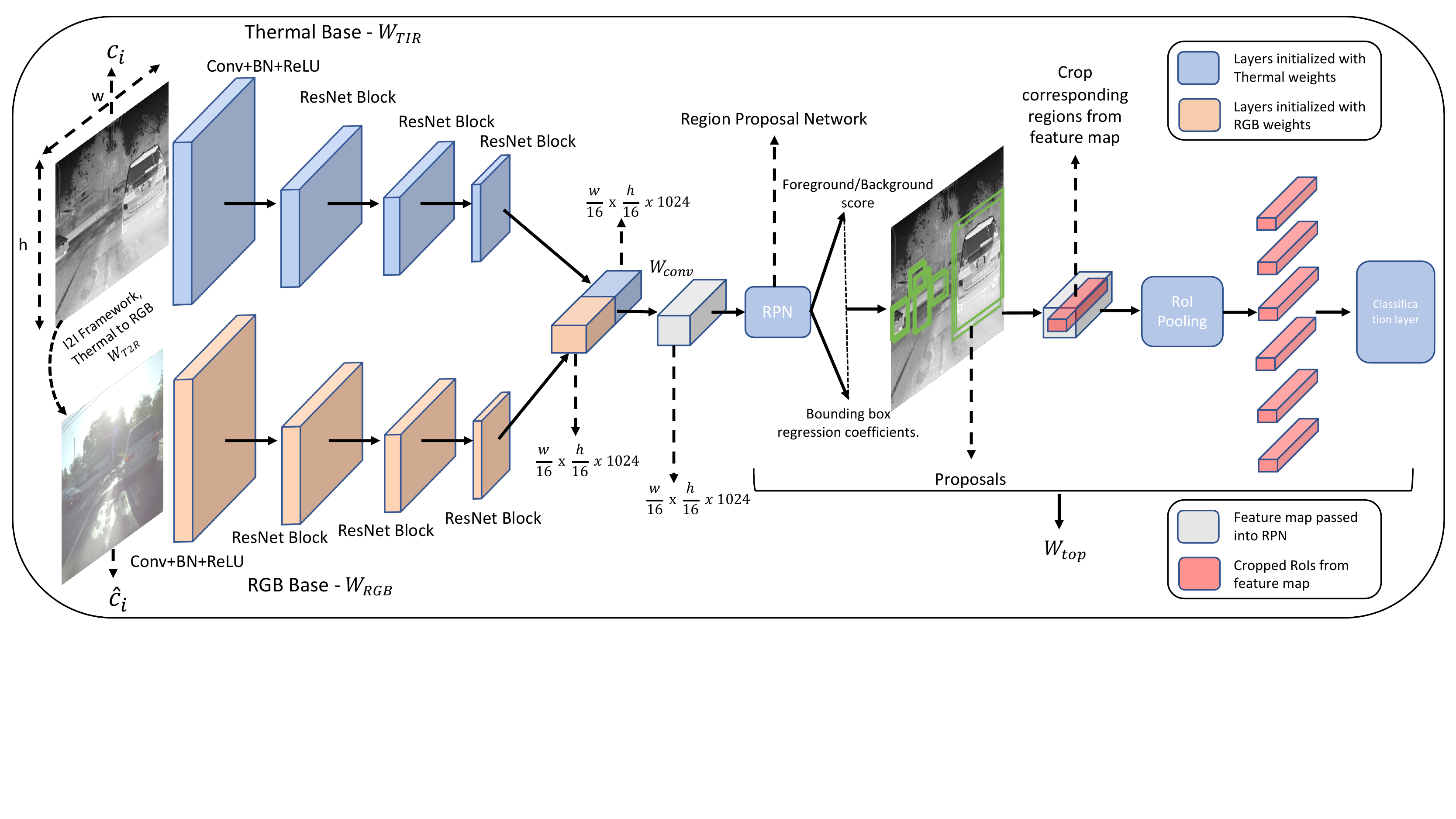}
\end{center}
\vspace{-7em}
   \caption{Adaptation of proposed Mutli-modal framework for Faster-RCNN} \label{framework}
\end{figure*}

\section{Related Work} \label{relatedwork}
Detection and classification of objects in the thermal imagery has been an active area of research in computer vision \cite{2007Zin}\cite{2013target_detection}\cite{MovingObject2015}\cite{kaist}, especially in the context of military and surveillance\cite{multichannelCNN}. There has been a significant amount of work on classifying and detecting people and objects in thermal imagery using standard computer vision and machine learning models, even before deep learning became popular. Bertozzi \textit{et al.} \cite{bertozzi} proposed a probabilistic template-based approach for pedestrian detection in far infrared (IR) images. They divided their algorithm into three parts: candidate generation, candidate filtering and validation of candidates. One main weakness of this approach is that it assumes the human is hotter than the background which may not be the case in many real-world scenarios. Davis \textit{et al.} \cite{davis-twostageapproach} proposed a two-stage template-based method to detect people in widely varying thermal imagery. To locate the potential person locations, a fast screening procedure is used with a generalized template and then an AdaBoost ensemble classifier is used to test the hypothesized person locations. Kai \textit{et al.} \cite{Jngling2009FeatureBP} proposed a local feature-based pedestrian detector on thermal data. They used a combination of multiple cues to find interest points in the images and used SURF \cite{surf} as features to describe these points. A codebook is then constructed to locate the object center. The challenge of this detector is whether a high performance can be achieved when local features are not obvious.

While these efforts have shown good performance for IR image classification and detection tasks over a small number of objects, they have been outperformed in recent years by deep learning models that enable more descriptive features to be learned. With the increase in popularity of deep neural networks, several methods have been proposed for applying deep learning methods to thermal images. Peng \textit{et al.} \cite{pengnirfacenet} proposed a Convolutional Neural Network (CNN) for face identification in near IR images. Their CNN is a modification of GoogLeNet but has a more compact structure. Lee \textit{et al.} \cite{LEE2016261} designed a lightweight CNN consisting of two convolutional layers and two subsampling layers for recognizing unsafe behaviors of pedestrians using thermal images captured from moving vehicles at night. They combined their lightweight CNN with a boosted random forest classifier. Chevalier \textit{et al.} \cite{chevalier:hal-01332061} proposed LR-CNN for automatic target recognition which is a deep architecture designed for classification of low-resolution images with strong semantic content. Rodger \textit{et al.} \cite{lwircnn} developed a CNN trained on short-to-midrange high resolution IR images containing six object classes (person, land vehicle, helicopter, aeroplane, unmanned aerial vehicle and false alarm) using an LWIR sensor. This network was successful at classifying other short to mid-range objects in unseen images, although it struggled to generalize to long range targets. Abbott \textit{et al.} \cite{Abbott2017} used a transfer learning approach with the YOLO \cite{yolo} framework to train a network on high-resolution thermal imagery for classification of pedestrians and vehicles in low-resolution thermal images. Berg \textit{et al.} \cite{2015BergRails}\cite{amanda_phd} proposed an anomaly-based obstacle detection method using a train-mounted thermal camera. Leykin \textit{et al.} \cite{leykin} proposed a fusion tracker and pedestrian classifier for multispectral pedestrian detection. Proposals for performing detection are generated using background subtraction and evaluated using periodic gait analysis. 

Among efforts that use a multimodal approach, Wagner \textit{et al.} \cite{wagner} applied Aggregated Channel Features (ACF) and Boosted Decision trees (BDT) for proposal generation and classified these proposals with a CNN, which fuses Visual and IR information.  Choi \textit{et al.} \cite{choietal} uses two separate region proposal networks for both Visual and IR images and evaluates the proposals generated by both the networks with Support Vector Regression on fused deep features. The efforts closest to our work are that of Konig \textit{et al.} \cite{daniel} and Liu \textit{et al.} \cite{Liu2016MultispectralDN}, both of which propose a multi-modal framework that combines RGB and thermal information in a Faster-RCNN architecture by posing it as a convolutional network fusion problem. However, all of these multimodal efforts assume the availability of a dataset with paired training examples from the visual and thermal domain. On the other hand, our work assumes only the presence of thermal imagery and seeks to leverage the use of publicly available RGB datasets (which may not be paired with the thermal dataset) to obtain significant improvement in thermal object detection performance.


\section{Methodology} \label{method}

Our overall proposed methodology for `pseudo multi-modal' object detection for thermal images is summarized in Figure \ref{framework}. The key idea of our methodology is to borrow knowledge from data-rich domains such as visual (RGB) without the explicit need for a paired multimodal dataset. We achieve this objective by leveraging the success of recent image-to-image translation methods \cite{CycleGAN2017, UNIT} to automatically generate a pseudo-RGB image from a given thermal image, and then propose a multimodal Faster R-CNN architecture to achieve our objective. Image-to-Image translation models aim to learn the visual mapping between a source domain and target domain. Learning this mapping becomes challenging when there are no paired images in source and target domains. Recently, there have been noteworthy efforts on addressing this problem using unpaired images \cite{CycleGAN2017}\cite{dualgan2017}\cite{stargan2017}\cite{UNIT}\cite{taigman2017unsupervised}\cite{instagan}\cite{drit2018}. While one could use any unsupervised image-to-image translation framework in our overall methodology, we use  CycleGAN\cite{CycleGAN2017} and UNIT\cite{UNIT} as I2I frameworks of choice in our work, owing to their wide use and popularity. We begin our discussion with the I2I translation frameworks used in this work.
\vspace{-1em}
\paragraph{Unpaired Image-to-Image Translation:} CycleGAN \cite{CycleGAN2017} is a popular unpaired image-to-image translation framework that aims to learn the mapping functions $F:\mathcal{X \rightarrow Y}$ and $G:\mathcal{Y \rightarrow X}$ where $\mathcal{X}$ and $\mathcal{Y}$ are source and target domains respectively. It maps the images onto two separate latent spaces and employs two generators $\mathcal{G_{X \rightarrow Y}}, \mathcal{G_{Y \rightarrow X}}$ and two discriminators $\mathcal{D_X}, \mathcal{D_Y}$. The generator $\mathcal{G_{X \rightarrow Y}}$ attempts to generate images $\hat{y}_i$ that look similar to images from domain $\mathcal{Y}$, while $\mathcal{D}_y$ aims to distinguish between the translated samples $\hat{y}_i$ and real samples $y_i$. This condition is enforced using an adversarial loss. To reduce the space of possible mapping functions, a cycle-consistency constraint is also enforced, such that a source-domain image $x_i$ when transformed into target domain ($\hat{y}_i$) and re-transformed back to source domain ($\hat{x}_i$) will ensure in $\hat{x}_i$ and $x_i$ will belong to the same distribution. For more details, please see \cite{CycleGAN2017}.

Unlike CycleGAN \cite{CycleGAN2017}, UNIT \cite{UNIT} tackles the unpaired image-to-image translation problem assuming a shared latent space between both the domains. It learns the joint distribution of images in different domains using the marginal distribution in individual domains. The framework is based on variational autoencoders $\mathcal{\text{VAE}_{\text{1}}}, \mathcal{\text{VAE}_{\text{2}}}$ and generative adversarial networks $\mathcal{\text{GAN}_{\text{1}}}, \mathcal{\text{GAN}_{\text{2}}}$ with a total of six sub-networks including two image encoders $\mathcal{E}_1, \mathcal{E}_2$, two image generators $\mathcal{G}_1, \mathcal{G}_2$ and two adversarial discriminators $\mathcal{D}_1, \mathcal{D}_2$. Since they assume a shared latent space between the two domains, a weight sharing constraint is enforced to relate the two VAEs. Specifically, weight sharing is done between the last few layers of encoders $\mathcal{E}_1, \mathcal{E}_2$ that are responsible for higher level representations of the input images in the two domains and the first few layers of the image generators $\mathcal{G}_1, \mathcal{G}_2$ responsible for decoding the high-level representations for reconstructing the input images. The learning problems of $\mathcal{\text{VAE}_{\text{1}}}, \mathcal{\text{VAE}_{\text{2}}}, \mathcal{\text{GAN}_{\text{1}}}, \mathcal{\text{GAN}_{\text{2}}}$ for image reconstruction, image translation and cyclic reconstruction are jointly solved. For more information, please see \cite{UNIT}.

In case of both CycleGAN and UNIT, the trained model provides two generators which perform the translation between source and target domains. In our case, we use the generator which performs the Thermal-to-RGB translation, which is given by  $G:\mathcal{X \rightarrow Y}$ in case of a CycleGAN and $G_1$ in case of UNIT (we used Thermal as the source domain, and RGB as the target domain while training these models). We refer to the parameters of these generators as $W_{T2R}$ in our methodology.

\begin{algorithm}[h]
\SetAlgoLined
\textbf{Input:} Thermal image training data: $\{(c_i, y_i)\}_{i=1}^m$; Generator of I2I framework: $W_{T2R}$; Pre-trained RGB base network: $W_{RGB}$; Pre-trained thermal base network: $W_{TIR}$, Pre-trained thermal top network $W_{top}$; Randomly initialised 1x1 conv weights: $W_{conv}$; Number of epochs: $num\_epochs$; Loss function: $\mathcal{L(.)}$\\
\textbf{Output:} Trained MMTOD model, $\mathcal{F(.)}$\\
\For{$num\_epochs$}{
    \For{$c_{i}, i= 1, \cdots, m$}{
            
            Generate a pseudo-RGB image $\hat{c}_i$ using $W_{T2R}$. \\ 
            Generate feature maps by passing $c_{i}$ and $\hat{c}_{i}$ to base networks $W_{TIR}$ and $W_{RGB}$ respectively\\ 
            Stack the feature maps and use $W_{conv}$ to get $1 \times 1$ conv output  \\
            Pass the 1x1 conv output to $W_{top}$ \\
            Update weights: $W_{RGB}, W_{TIR}, W_{top}, W_{conv}, W_{T2R}$ by minimizing $\mathcal{L}$ of the object detection framework.
        
    }
}
\caption{MMTOD: Multi-modal Thermal Object Detection Methodology}
\label{alg_mmtod}
\end{algorithm}


\paragraph{Pseudo Multi-modal Object Detection:} As shown in Figure \ref{framework}, our object detection framework is a multi-modal architecture consisting of two branches, one for the thermal image input and the other for the RGB input. 

Each branch is initialized with a model pre-trained on images from that domain (specific details of implementation are discussed in Section \ref{exp_section}). To avoid the need for paired training examples from two modalities but yet use a multi-modal approach, we use an image-to-image (I2I) translation network in our framework. During the course of training, for every thermal image input, we generate a pseudo-RGB using $W_{T2R}$ and pass the pseudo-RGB and  Thermal to the input branches (parametrized by $W_{RGB}$ and $W_{TIR}$ respectively). Outputs from these branches are stacked and passed through a $1 \times 1$ convolution ($W_{conv}$) to learn to combine these features appropriately for the given task. The output of this $1 \times 1$ convolution is directly passed into the rest of the Faster-RCNN network (denoted by $W_{top}$). We use the same Region Proposal Network (RPN) loss as used in Faster-RCNN, given as follows: 
\begin{equation*}
\label{eq:loss}
    L(\{p_i\}, \{t_i\}) = \frac{1}{N_{cls}} \sum_iL_{cls}(p_i, p^{*}_i) + \lambda \frac{1}{N_reg} \sum_i p^{*}_{i}L_{reg}(t_i, t^{*}_{i})  
\end{equation*}
\noindent where $i$ is the index of an anchor, $p_i$ is the predicted probability of anchor $i$ being an object, $p^{*}_i$ is the ground truth, $t_i$ represents the coordinates of the predicted bounding box, $t^{*}_i$ represents the ground truth bounding box coordinates, $L$ is log loss, $R$ is the robust loss function (smooth L$_1$) as defined in \cite{fastrcnn}, and $\lambda$ is a hyperparameter. We use the same multi-task classification and regression loss as used in Fast-RCNN \cite{fastrcnn} at the end of the network. 

While the use of existing I2I models allow easy adoption of the proposed methodology, the images generated from such I2I frameworks for thermal-to-RGB translation are perceptually far from natural RGB domain images (like MS-COCO\cite{coco} and PASCAL-VOC \cite{pascal-voc-2007}), as shown in Figure \ref{trans_examples}. Therefore, during the training phase of our multi-modal framework, in order to learn to combine the RGB and thermal features in a way that helps improve detection, we also update the weights of the I2I generator $W_{T2R}$. This helps learn a better representation of the pseudo-RGB image for borrowing relevant features from the RGB-domain, which we found to be key in improving detection in the thermal domain. The proposed methodology provides a fairly simple strategy to improve object detection in the thermal domain. We refer to the proposed methodology as MMTOD (Multimodal Thermal Object Detection) hereafter. Our algorithm for training is summarized in Algorithm \ref{alg_mmtod}. More details on the implementation of our methodology are provided in Section \ref{exp_section}.


\section{Experiments} \label{exp_section}
\subsection{Datasets and Experimental Setup}

\begin{figure}[H]
\centering
   \includegraphics[width=\hsize]{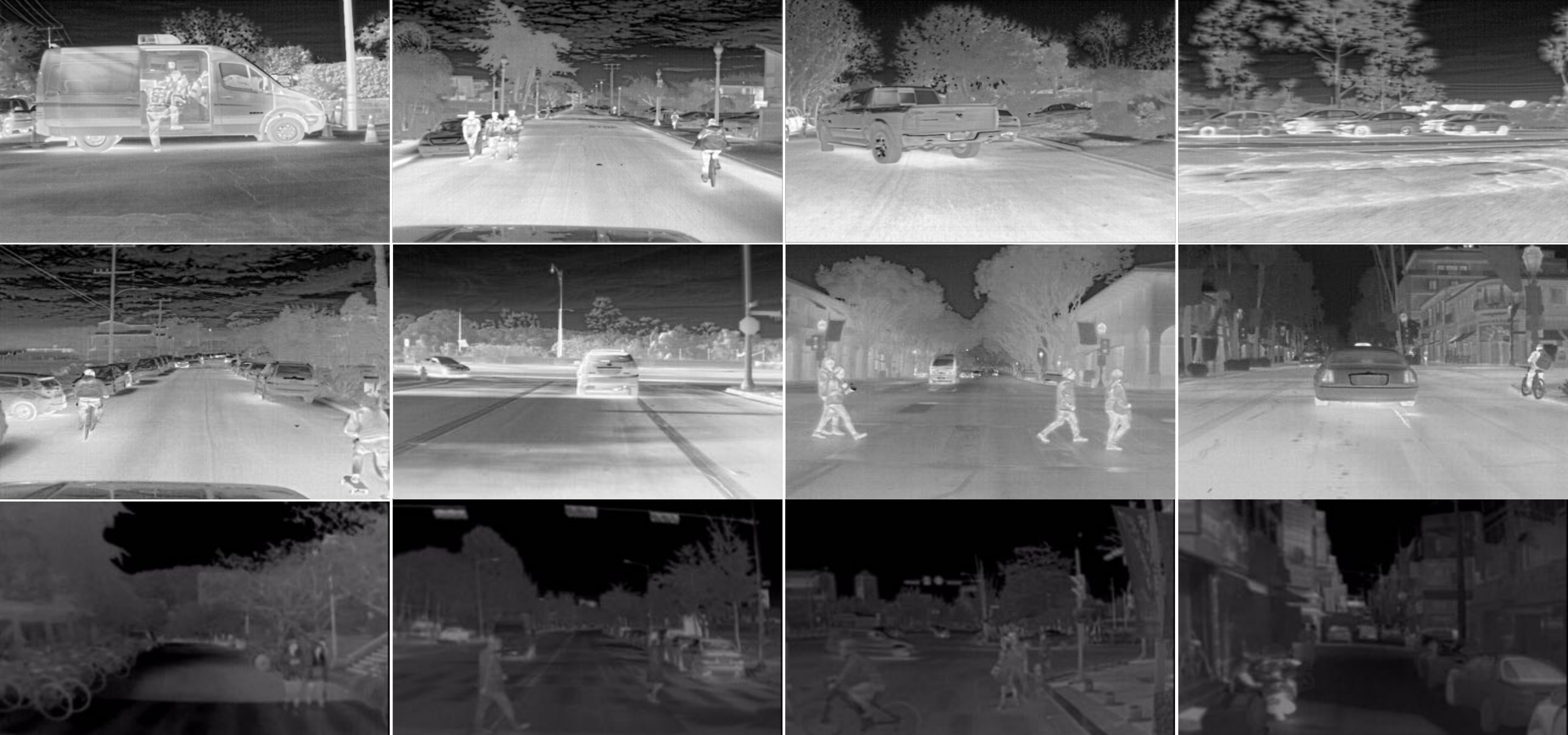}
   \caption{\textit{Row 1} \& \textit{Row 2}: Example images from FLIR \cite{flir} ADAS dataset, \textit{Row 3}: Example Images from KAIST \cite{kaist} dataset} \label{example_images}
  \vspace{-0.9em}

\end{figure}

\paragraph{Datasets:} We use the recently released FLIR ADAS \cite{flir} dataset and the KAIST Multispectral Pedestrian dataset \cite{kaist} for our experimental studies. FLIR ADAS \cite{flir} consists of a total of 9,214 images with bounding box annotations, where each image is of $640 \times 512$ resolution and is captured using a FLIR Tau2 camera. 60\% of the images are collected during daytime and the remaining 40\% are collected during night. While the dataset provides both RGB and thermal domain images (not paired though), we use only the thermal images from the dataset in our experiments (as required by our method). For all the experiments, we use the training and test splits as provided in the dataset benchmark, which contains the person (22,372 instances), car (41,260 instances), and bicycle (3,986 instances) categories. Some example images from the dataset are shown in Figure \ref{example_images}. 

The KAIST Multispectral pedestrian benchmark dataset \cite{kaist} contains around 95,000 8-bit day and night images (consisting of only the Person class). These images are collected using a FLIR A35 microbolometer LWIR camera with a resolution of $320 \times 256$ pixels. The images are then upsampled to $640 \times 512$ in the dataset. Sample images from the dataset are shown in Figure \ref{example_images}. Though the KAIST dataset comes with fully aligned RGB and Thermal, we choose not to use the RGB images as our goal to improve the detection in the absence of paired training data. 

Our methodology relies on using publicly available large-scale RGB datasets to improve thermal object detection performance. For this purpose, we use RGB datasets with the same classes as in the aforementioned thermal image datasets. In particular, we perform experiments using two popular RGB datasets namely, PASCAL VOC \cite{pascal-voc-2007} and MS-COCO \cite{coco}. In each experiment, we pre-train an object detector on either of these datasets and use these parameters to initialise the RGB branch of our multimodal framework. We also compare the performance of these two initializations in our experiments. In case of thermal image datasets, an end-to-end object detector is first trained on the dataset and used to initialize the thermal branch of our framework. We use mean Average Precision (mAP) as the performance metric, as is common for the object detection task. 

\vspace{-4pt}
\paragraph{Baseline:} A Faster-RCNN trained in a fully supervised manner on the thermal images from the training set is used as the baseline method for the respective experiments in our studies. 
We followed the original paper \cite{fasterrcnn} for all the hyperparameters, unless specified otherwise. The FLIR ADAS dataset \cite{flir} also provides a benchmark test mAP (at IoU of 0.5) of 0.58 using the more recent  RefineDetect-512 \cite{refineDet} model. We show that we beat this benchmark using our improved multi-modal Faster-RCNN model.

\vspace{-4pt}

\paragraph{Image-to-Image Translation (IR-to-RGB):} For our experiments, we train two CycleGAN models: one for FLIR $\leftrightarrow$  RGB which uses thermal images from FLIR \cite{flir} and RGB images from PASCAL VOC \cite{pascal-voc-2007}, and another for KAIST $\leftrightarrow$ RGB which uses thermal images from KAIST \cite{kaist} and RGB images from PASCAL VOC \cite{pascal-voc-2007}. We use an initial learning rate of 1e-5 for the first 20 epochs, which is decayed to zero over the next 20 epochs. The identity mapping is set to zero, i.e., the identity loss and the reconstruction loss are given equal weightage. The other hyperparameters of training are as described in \cite{CycleGAN2017}. For training of the UNIT framework, all the hyperparameters are used as stated in the original paper, without any alterations. Since UNIT takes a long time to train (7 to 8 days on an NVIDIA P-100 GPU), we trained it only for FLIR $\leftrightarrow$  RGB, so the experiments on KAIST are performed using CycleGAN only. Our variants are hence referred to as MMTOD-CG (when I2I is CycleGAN) and MMTOD-UNIT (when I2I is UNIT) in the remainder of the text.

We use the same metrics as mentioned in CycleGAN \cite{CycleGAN2017} and UNIT \cite{UNIT} papers for evaluating the quality of translation. In an attempt to improve the quality of generated images in CycleGAN \cite{CycleGAN2017}, we tried adding feature losses in addition to cycle consistency loss and adversarial loss. However, this did not improve the thermal to visual RGB translation performance. We hence chose to finally use the same loss as mentioned in \cite{CycleGAN2017}.

\paragraph{Training our Multi-modal Faster-RCNN:} Our overall architecture (as in Figure \ref{framework}) is initialized with pre-trained RGB and Thermal detectors as described in Section \ref{method}. Since our objective is to improve detection in thermal domain, the region proposal network (RPN) is initialized with weights pre-trained on thermal images. The model is then trained on the same set of images on which the thermal detector was previously pre-trained. The I2I framework generates a pseudo-RGB image corresponding to the input thermal image. The thermal image and the corresponding pseudo-RGB image are passed through the branches of the multi-modal framework to obtain two feature maps of 1024 dimension each, as shown in figure \ref{framework}. These two feature maps are stacked back-to-back and passed through a $1 \times 1$ convolution, which is then passed as input to the Region Proposal Network (RPN). RPN produces the promising Regions of Interest (RoIs) that are likely to contain a foreground object. These regions are then cropped out of the feature map and passed into a classification layer which learns to classify the objects in each ROI. Note that as mentioned in Section \ref{method}, during the training of the MMTOD framework, the weights of the I2I framework are also updated which allows it to learn a better representation of the translated image for improved object detection in thermal domain. We adapted the Faster-RCNN code provided at \cite{jjfaster2rcnn} for our purpose. The code for the CycleGAN and UNIT was taken from their respective official code releases\cite{CycleGAN2017}\cite{isola2017image}\cite{UNIT}. Our code will be made publicly available for further clarifications. 
\paragraph{Experimental Setup:} To evaluate the performance of the proposed multi-modal framework, the following experiments are carried out:
\vspace{-1em}
\begin{itemize}
    \item MMTOD-CG with RGB branch initialized by PASCAL-VOC pre-trained detector, thermal branch initialized by FLIR ADAS pre-trained detector
    \vspace{-0.85em}
    \item MMTOD-UNIT with RGB branch initialized by PASCAL-VOC pre-trained detector, thermal branch initialized by FLIR ADAS pre-trained detector
    \vspace{-0.85em}
    \item MMTOD-CG with RGB branch initialized by MS-COCO pre-trained detector, thermal branch initialized by FLIR ADAS pre-trained detector
      \vspace{-0.85em}
    \item MMTOD-UNIT with RGB branch initialized by MS-COCO pre-trained detector, thermal branch initialized by FLIR ADAS pre-trained detector
      \vspace{-0.85em}
    \item MMTOD-CG with RGB branch initialized by PASCAL-VOC pre-trained detector, thermal branch initialized by KAIST pre-trained detector
      \vspace{-0.85em}
    \item MMTOD-CG with RGB branch initialized by COCO pre-trained detector, thermal branch initialized by KAIST pre-trained detector
\end{itemize}
\begin{figure*}
\begin{center}
\includegraphics[width=16.5cm, height=5cm]{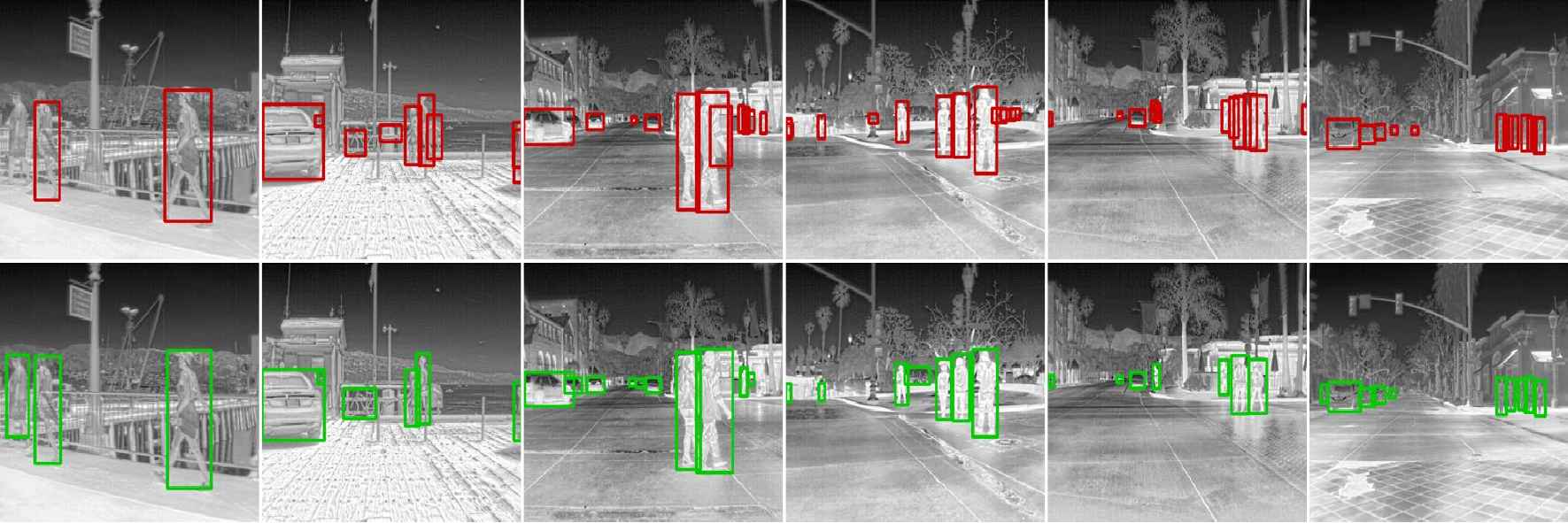}
\end{center}
   \caption{Qualitative results of detections on the FLIR ADAS dataset. \textit{Row 1:} Baseline \textit{Row 2:} MMTOD}
   \label{rresults_flir}
\end{figure*}

\begin{figure*}
\begin{center}
\includegraphics[width=16.5cm, height=5cm]{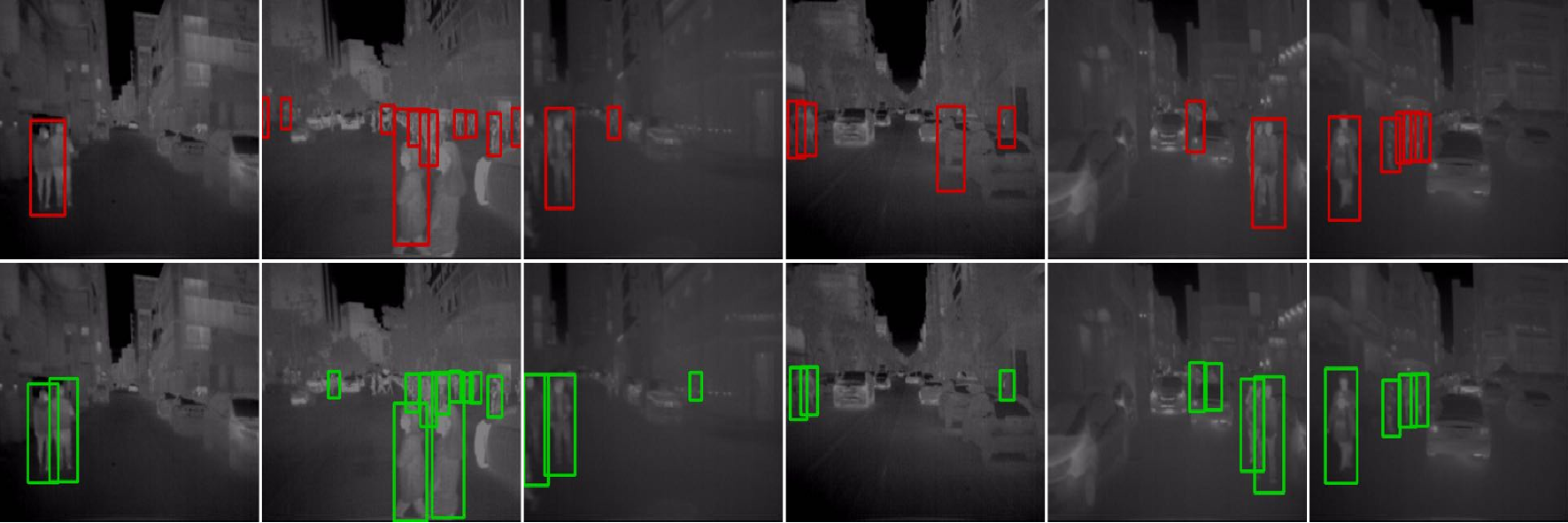}
\end{center}
  \caption{Qualitative results of detections on the KAIST. \textit{Row 1:} Baseline. \textit{Row 2:} MMTOD}
  \label{results_kaist}
\end{figure*}

\subsection{Results}
\begin{figure}[H]
\centering
   \includegraphics[width=\hsize]{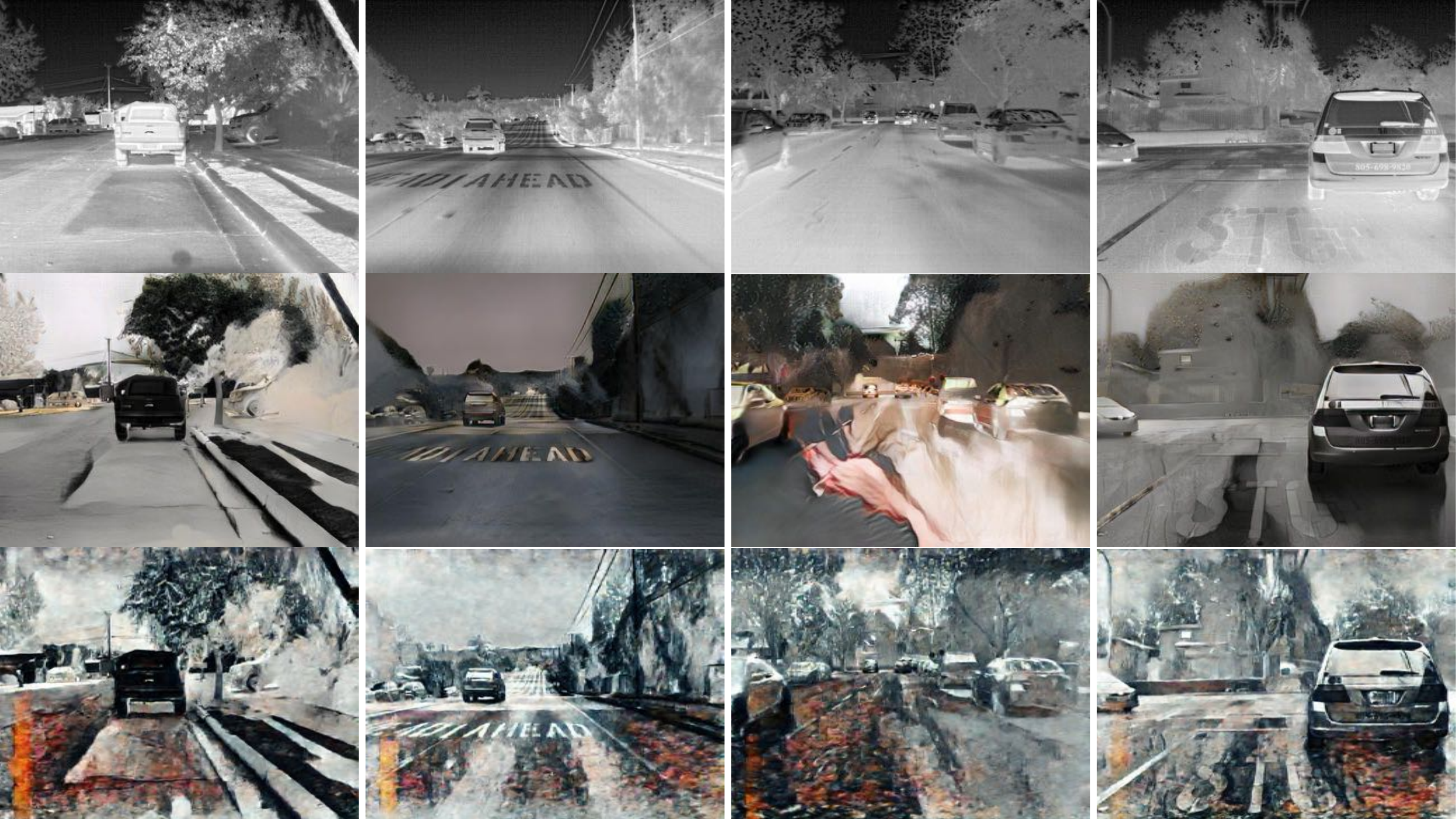}
   \caption{\textit{Row 1:} Thermal images from FLIR ADAS\cite{flir} dataset; \textit{Row 2:} Translations generated using  UNIT\cite{UNIT}; \textit{Row 3:} Translations generated using CycleGAN\cite{CycleGAN2017}.} \label{trans_examples}
\end{figure}

\paragraph{IR-to-RGB Translation Results:} Figure \ref{trans_examples} shows the results of CycleGAN and UNIT trained for Thermal $\leftrightarrow$ RGB translation. As mentioned in Section \ref{method}, the generated pseudo-RGB images are perceptually far from natural domain images. This can be attributed to the fact that the domain shift between RGB and Thermal domains is relatively high compared to other domains. In addition, RGB images have both chrominance and luminance information, while thermal images just have the luminance part which makes estimating the chrominance for RGB images a difficult task. However, we show that using our method, these generated images add value to the detection methodology.

\paragraph{Thermal Object Detection Results:}
Tables \ref{tbl:flir_each_class_comparision} and \ref{tbl:kaist_person_comparision} show the comparison of AP for each class and the mAP of our framework against the baseline detector when trained on FLIR ADAS and KAIST datasets respectively. (Note that the KAIST dataset has only one class, the Person.) We observe that in all the experiments, our framework outperforms the baseline network across all the classes.

\begin{table}[H]
\tabcolsep=1.5pt
\begin{tabular}{cccccc}
\hline
                                                &               & \multicolumn{3}{c}{\textit{AP across each class}} &       \\ \cline{3-5}
\multicolumn{1}{l}{\textit{Method}}                      &               & \textit{Bicycle}    & \textit{Person}    & \textit{Car}    & \textit{mAP}   \\ \hline
\multicolumn{1}{l}{Baseline}                    &               & 39.66         & 54.69        & 67.57     & 53.97 \\ \hline

\multicolumn{1}{c|}{Framework}                 & RGB Branch &               &              &           &       \\ \cline{1-2}
\multicolumn{1}{c|}{\multirow{2}{*}{MMTOD-UNIT}}      & MSCOCO        & 49.43         & \textbf{64.47}        & \textbf{70.72}     & \textbf{61.54} \\
\multicolumn{1}{c|}{}                           & Pascal VOC    & 45.81         & 59.45        & 70.42     & 58.56 \\ \hline
\multicolumn{1}{c|}{\multirow{2}{*}{MMTOD-CG}} & MSCOCO        & 50.26         & 63.31        & 70.63     & 61.40 \\
\multicolumn{1}{c|}{}                           & Pascal VOC    & 43.96         & 57.51        & 69.85     & 57.11 \\ \hline
\end{tabular}
 \caption{Performance comparison of proposed methodology against baseline on FLIR \cite{flir}}
 \label{tbl:flir_each_class_comparision}
\end{table}

\begin{table}[H]
\centering
\begin{tabular}{lcc}
\hline
\textit{Method}                                          &            & \textit{mAP} \\ \hline
Baseline                                        &            & 49.39     \\ \hline
\multicolumn{1}{c|}{Framework}             & RGB Branch &           \\ \cline{1-2}
\multicolumn{1}{l|}{\multirow{2}{*}{MMTOD-CG}} & MS-COCO    & \textbf{53.56}      \\
\multicolumn{1}{l|}{}                           & Pascal VOC & 52.26     \\ \hline
\end{tabular}
\caption{Performance comparison of proposed methodology against baseline on KAIST \cite{kaist}}
\label{tbl:kaist_person_comparision}
\end{table}

In case of FLIR, we observe that initializing the RGB branch with MS-COCO obtains better results than those with PASCAL-VOC. This can be attributed to the fact that MS-COCO has more instances of car, bicycle, and person as compared to PASCAL VOC. Also, experimental results show that employing UNIT as the I2I framework achieves better performance than CycleGAN. Our framework with MS-COCO initialization and UNIT for I2I translation results in an increase in mAP by at least 7 points. In particular, as mentioned earlier, the FLIR ADAS dataset provides a benchmark test mAP (at IoU of 0.5) of 0.58 using the more recent RefineDetect-512 \cite{refineDet} model. Our method outperforms the benchmark despite using a relatively older object detection model such as the Faster-RCNN. 

As shown in Table \ref{tbl:kaist_person_comparision}, our improved performance on the KAIST dataset shows that although this dataset has more examples of the 'Person' category than the RGB dataset used such as PASCAL-VOC, our framework still improves upon the performance of the baseline method. This allows us to infer that the proposed framework can be used in tandem with any region-CNN based object detection method to improve the performance of object detection in thermal images. On average our framework takes 0.11s to make detections on a single image, while the baseline framework takes 0.08s. Our future directions of work include improving the efficiency of our framework while extending the methodology to other object detection pipelines such as YOLO and SSD.

\section{Discussion and Ablation Studies} 
\label{discussion}
\paragraph{Learning with limited examples:} We also conducted studies to understand the capability of our methodology when there are limited samples in the thermal domain. Our experiments on the FLIR ADAS dataset showed that our framework outperforms the current state-of-the-art detection performance using only half the training examples. Moreover, our experiments show that using only a quarter of the training examples, our framework outperforms the baseline on the full training set. Table \ref{tbl:stats_datasets} presents the statistics of the dataset used for this experiment. Note that the test set used in these experiments is still the same as originally provided in the dataset. 

\begin{table}[H]
  \centering
\begin{tabular}{@{}llll@{}}
\toprule
          & \multicolumn{3}{c}{Number of Instances} \\ \cmidrule(l){2-4} 
Dataset    & Car         & Person      & Bicycle     \\ \midrule
FLIR       & 41,260      & 22,372      & 3,986       \\
FLIR (1/2) & 20,708      & 11,365      & 2,709       \\
FLIR (1/4) & 10,448      & 5,863       & 974         \\ \bottomrule
\end{tabular}
\caption{Statistics of the datasets we used for our experiments.}
  \label{tbl:stats_datasets}
\vspace{-1em}
\end{table}
We perform the same set of experiments (as discussed in Section \ref{exp_section}) on FLIR(1/2) and FLIR(1/4) datasets. Tables \ref{tbl:flir_half_each_class_comparision} and \ref{tbl:flir_quarter_each_class_comparision} present the results.

\begin{table}[H]
\resizebox{0.48\textwidth}{!}{%
    \tabcolsep=1.5pt
\begin{tabular}{cccccc}
\hline
                                                                                           &            & \multicolumn{3}{c}{\textit{AP across each class}} &       \\ \cline{3-5}
\multicolumn{1}{l}{\textit{Method}}                                                                 &            & \textit{Bicycle}    & \textit{Person}    & \textit{Car}    & \textit{mAP}   \\ \hline
\multicolumn{1}{l}{Baseline (FLIR)}                    &               & 39.66         & 54.69        & 67.57     & 53.97 \\ \hline
\multicolumn{1}{l}{Baseline (FLIR-1/2)}                                                               &            & 34.41         & 51.88        & 65.04     & 50.44 \\ \hline

\multicolumn{1}{c|}{Framework}                                                         & RGB Branch &               &              &           &       \\ \cline{1-2}
\multicolumn{1}{c|}{\multirow{2}{*}{MMTOD-UNIT}}                                                 & MSCOCO     & 49.84         & \textbf{59.76}        & \textbf{70.14}     & \textbf{59.91} \\
\multicolumn{1}{c|}{}                                                                      & Pascal VOC & 45.53         & 57.77        & 69.86     & 57.72 \\ \hline
\multicolumn{1}{c|}{\multirow{2}{*}{\begin{tabular}[c]{@{}c@{}}MMTOD-CG\end{tabular}}} & MSCOCO     & 50.19         & 58.08        & 69.77     & 59.35 \\
\multicolumn{1}{c|}{}                                                                      & Pascal VOC & 40.17         & 54.67        & 67.62     & 54.15 \\ \hline
\end{tabular}}
    \caption{Performance comparison of proposed methodology against baseline on FLIR (1/2)}
    \label{tbl:flir_half_each_class_comparision}

\end{table}

\begin{table}[H]
\resizebox{0.48\textwidth}{!}{%
        \tabcolsep=1.5pt
\begin{tabular}{cccccc}
\hline
                                                &            & \multicolumn{3}{c}{\textit{AP across each class}} &       \\ \cline{3-5}
\multicolumn{1}{l}{\textit{Method}}                      &            & \textit{Bicycle}       & \textit{Person}      & \textit{Car}        & \textit{mAP}   \\ \hline
\multicolumn{1}{l}{Baseline(FLIR)}                    &               & 39.66         & 54.69        & 67.57     & 53.97 \\ \hline
\multicolumn{1}{l}{Baseline(FLIR-1/4)}               &            & 33.35         & 49.18           & 60.84     & 47.79 \\ \hline

\multicolumn{1}{c|}{Framework}              & RGB Branch &               &             &            &       \\ \cline{1-2}
\multicolumn{1}{c|}{\multirow{2}{*}{MMTOD-UNIT}}      & MSCOCO     & \textbf{44.24}         & \textbf{57.76}       & \textbf{69.77}      & \textbf{57.26} \\
\multicolumn{1}{c|}{}                           & Pascal VOC & 35.23         & 54.71       & 67.83      & 52.59 \\ \hline
\multicolumn{1}{c|}{\multirow{2}{*}{MMTOD-CG}} & MSCOCO     & 41.29         & 57.08       & 69.10      & 55.82 \\
\multicolumn{1}{c|}{}                           & Pascal VOC & 35.02         & 51.62       & 66.09      & 50.91 \\ \hline
\end{tabular}}
 \caption{Performance comparison of proposed methodology against baseline on FLIR (1/4)}
    \label{tbl:flir_quarter_each_class_comparision}

\end{table}
\vspace{-1em}
Table \ref{tbl:flir_half_each_class_comparision} shows the baselines for training the Faster-RCNN on the complete FLIR training dataset as well as FLIR (1/2). We observe that both MMTOD-UNIT and MMTOD-CG trained on FLIR(1/2) outperform both the baselines, even when Faster-RCNN is trained on the entire training set. 

Similarly, Table \ref{tbl:flir_quarter_each_class_comparision} shows the baselines for training the Faster-RCNN on the complete FLIR training dataset as well as FLIR (1/4). Once again, we observe that both MMTOD-UNIT and MMTOD-CG trained on FLIR(1/4) outperform both the baselines, even when Faster-RCNN is trained on the entire training set. In other words, the MMTOD framework requires only a quarter of the thermal training set to surpass the baseline accuracy achieved using the full training set. The results clearly demonstrate the proposed framework's ability to learn from fewer examples. This shows that our framework effectively borrows features from the RGB domain that help improve detection in the thermal domain. This is especially useful in the context of thermal and IR images, where there is a dearth of publicly available large-scale datasets. 

\vspace{-1em}

\paragraph{Effect of Image Resolution:} To understand the effect of image resolution on object detection performance, we repeated the above experiments were conducted using subsampled images of the FLIR ADAS dataset. Table \ref{tbl:flir_400_exps} presents these results for $400 \times 400$ input images. We observe that our multi-modal framework improves the object detection performance significantly even in this case. Our future work will involve extending our work to images of even lower resolutions.
\begin{table}[H]
\resizebox{0.48\textwidth}{!}{%
        \tabcolsep=1.5pt
\begin{tabular}{cccccc}
\hline
                                                 &                  & \multicolumn{3}{c}{\textit{AP across each class}} &       \\ \cline{3-5}
\textit{Dataset}                                          & \textit{Method}           & \textit{Bicycle}     & \textit{Person}    & \textit{Car}    & \textit{mAP}   \\ \hline
\multicolumn{1}{c|}{\multirow{2}{*}{FLIR}}       & Baseline         & 29.25         & 43.13        & 58.83     & 43.74 \\
\multicolumn{1}{c|}{}                            & P-VOC + CycleGAN & \textbf{39.42}         & \textbf{52.75}        & \textbf{62.05}     & \textbf{51.41} \\ \cline{2-6} 
\multicolumn{1}{c|}{\multirow{2}{*}{FLIR (1/2)}} & Baseline         & 23.31         & 40.82        & 56.25     & 40.13 \\
\multicolumn{1}{c|}{}                            & P-VOC + CycleGAN & \textbf{33.32}         & \textbf{48.32}        & \textbf{60.87}     & \textbf{47.50} \\ \cline{2-6} 
\multicolumn{1}{c|}{\multirow{2}{*}{FLIR (1/4)}} & Baseline         & 18.81         & 35.42        & 52.82     & 35.68 \\
\multicolumn{1}{c|}{}                            & P-VOC + CycleGAN & \textbf{30.63}         & \textbf{45.45}        & \textbf{60.32}     & \textbf{45.47} \\ \hline
\end{tabular}}
 \caption{Performance comparison of proposed methodology against baseline on FLIR $400 \times 400$ images}
     \label{tbl:flir_400_exps}

\end{table}

\vspace{-2em}

\begin{figure}[H]
\centering
   \includegraphics[width=\hsize, height=5cm]{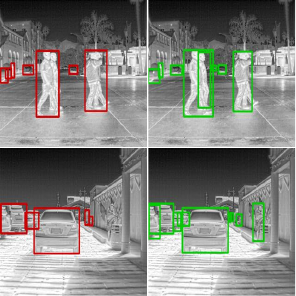}
  \caption{Some examples of missed detections, \textit{Red}: Predictions using MMTOD, \textit{Green}: Ground Truth} \label{missed_det}
\end{figure}

\paragraph{Missed Detections:} We tried to analyze the failure cases of the proposed methodology, by studying the missed detections. Some examples of these missed detections are shown in figure \ref{missed_det}. We infer that MMTOD finds object detection challenging when: (i) the objects are very small and located far from the camera; (ii) two objects are close to each other, and are detected as a single object; and (iii) there is heavy occlusion and crowd. Our future efforts will focus on addressing these challenges.

\vspace{-2em}

\section{Conclusion}
We propose a novel multi-modal framework to extend and improve upon any Region-CNN-based object detector in the thermal domain by borrowing features from the RGB domain, without the need of paired training examples. We evaluate the performance of our framework applied to a Faster-RCNN architecture in various settings including the FLIR ADAS and KAIST datasets. We demonstrate that our framework achieves better performance than the baseline, even when trained only on quarter of the thermal dataset. The results suggest that our framework provides a simple and straightforward strategy to improve the performance of object detection in thermal images. 

\section*{Acknowledgements}
This work was carried out as part of a CARS project supported by ANURAG, Defence Research and Development Organisation (DRDO), Government of India.

\newpage

\clearpage
{\small
\bibliographystyle{ieee}
\bibliography{paper.bbl}
}

\end{document}